# Self-Supervised In-Domain Representation Learning for Remote Sensing Image Scene Classification

Ali Ghanbarzadeh and Hossein Soleimani

**Abstract:** Transferring the ImageNet pre-trained weights to the various remote sensing tasks has produced acceptable results and reduced the need for labeled samples. However, the domain differences between ground imageries and remote sensing images cause the performance of such transfer learning to be limited. Recent research has demonstrated that self-supervised learning methods capture visual features that are more discriminative and transferable than the supervised ImageNet weights. We are motivated by these facts to pre-train the in-domain representations of remote sensing imagery using contrastive self-supervised learning and transfer the learned features to other related remote sensing datasets. Specifically, we used the SimSiam algorithm to pre-train the in-domain knowledge of remote sensing datasets and then transferred the obtained weights to the other scene classification datasets. Thus, we have obtained state-of-the-art results on five land cover classification datasets with varying numbers of classes and spatial resolutions. In addition, By conducting appropriate experiments, including feature pre-training using datasets with different attributes, we have identified the most influential factors that make a dataset a good choice for obtaining in-domain features. We have transferred the features obtained by pre-training SimSiam on remote sensing datasets to various downstream tasks and used them as initial weights for fine-tuning. Moreover, we have linearly evaluated the obtained representations in cases where the number of samples per class is limited. Our experiments have demonstrated that using a higher-resolution dataset during the self-supervised pre-training stage results in learning more discriminative and general representations.

**Keywords**: Transfer Learning, Deep Learning, Remote Sensing, Self-Supervised Learning, Representation Learning, Scene Classification

## 1. INTRODUCTION

Remote sensing imageries are acquired via imaging satellites, airplanes, etc. [1]. These devices are capable of monitoring various aspects of the earth's surface. Unlike natural images, which are captured using digital cameras and often contain a limited number of objects, remote sensing imageries can encompass vast geographical areas and hold numerous contents with varying dimensions and sizes. Remote sensing images, in contrast to ground images, are not object-centric. Therefore, they can be used for various applications, including land cover classification, road network extraction, disaster prevention, and monitoring [2, 3]. Only artificial intelligence and machine learning systems can process this volume of data. Fortunately, recent advances in computer vision have made it easier to process and analyze visual data points[4]. With the recent advancements in deep learning for computer vision applications, the supervised learning approaches for land cover classification in remote sensing images have performed exceptionally well. However, the main drawback of supervised learning is that it needs a tremendous amount of labeled samples. Providing this volume of remote-sensing images is very costly and time-consuming. In addition, it requires experts to annotate data points carefully. When obtaining large labeled datasets is exhaustive, the general solution is to transfer the learned weights from the ImageNet dataset to these tasks [1, 5-13]. While this transfer learning has produced acceptable results for remote sensing tasks, It has the following drawbacks:

1. If there are significant domain differences between the remote sensing and the ImageNet datasets, this type of transfer learning will fail. As a result, it can perform poorly in some cases, such as hyperspectral and multispectral images.
2. Transferring ImageNet pre-trained weights directly to non-RGB remote sensing datasets is impossible [3, 14-19].

Domain differences between the natural and remote sensing images stimulate researchers to find alternative solutions. To do so, some researchers used supervised or unsupervised methods to pre-trained models on remote sensing datasets. The learned weights are then transferred to other remote-sensing tasks[5, 19]. However, the disadvantage of the supervised pre-training is that it requires large in-domain labeled samples to learn general representations from remote sensing images. Self-supervised learning has emerged to overcome all of the previously mentioned drawbacks. It aims to learn effective representations of the input data without relying on human-provided labels**.** Recent advances in self-supervised learning have demonstrated that their pre-trained features have a higher transferability than ImageNet models. This branch of artificial intelligence is advantageous when acquiring labeled data is time-consuming and expensive, such as medical and satellite images[14]. Additionally, these methods are more practical in the real world because different sensors generate millions or even billions of data samples, and labeling them is unavoidably impossible. Recently, contrastive self-supervised learning [20] outperformed other feature learning methods. These methods significantly narrowed the gap between supervised and unsupervised approaches in representation learning [21]. Currently, the most effective contrastive self-supervised algorithms[22] employ data augmentation techniques to generate positive samples. In other words, they use data augmentation techniques such as image cropping, rotation, and so on to create multiple views of the same image. The objective function tries to bring positive samples as close to each other as possible in the feature space. In most of these methods, positive and negative pairs compete with each other [23]. Since these methods do not require labeled data, we can use a large amount of unlabeled data to learn the features in an unsupervised way and then transfer the weights to other remote sensing tasks.

Selecting an appropriate dataset for visual representation learning from remote sensing images, either supervised or self-supervised, is one of the influencing factors in learning high-generalizable features, which have huge effects on the performance of the final model on downstream tasks. In recent works, such as [6] and [17], researchers have determined the influencing factors on the datasets for pre-training features in a supervised manner from satellite images. Unlike supervised learning, which uses accurate human-provided labels as supervisory signals, self-supervised learning methods extract supervisory signals from the data itself. This difference makes it necessary to carefully investigate the vital factors that make the dataset an ideal option for self-supervised pre-training in remote sensing. One of our goals is to investigate the effect of the selected dataset for pre-training visual features from satellite images using the SimSiam algorithm. To achieve this goal, in the pre-training stage, we used datasets with different characteristics in terms of the number of samples, spatial resolution, and the number of classes. Our other goal is to investigate the generalizability of self-supervised learned features using the SimSiam for land cover classification. We have examined the transferability of the in-domain pre-trained weights by conducting extensive experiments. In the SimSiam algorithm, we used ResNet50 with ImageNet weights as the backbone. In this setting, we have pre-trained the features in a self-supervised manner on MLRSNet, PatternNet, and Resisc45 datasets. Finally, we have fine-tuned obtained models on the target datasets under different conditions, such as fine-tuning all layers and linear evaluation using a limited number of samples. The results demonstrate that by selecting a suitable medium-sized remote sensing dataset, we can pre-train features that produce the best results for various land-cover classification tasks. Our main contributions are as follows:

1. We have investigated the generalizability of the SimSiam algorithm for learning visual representations in remote sensing images by conducting detailed and exhaustive experiments on six land cover classification datasets with different characteristics.

2. During pre-training in-domain representations with the SimSiam algorithm, we used ImageNet weights as initial weights to reduce the need for training data.

3. By conducting detailed experiments, we have discussed the factors that make the dataset a good reference for self-supervised pre-training of features. The obtained results have demonstrated that the pre-training dataset should have a high spatial resolution.

The remainder of this paper is organized as follows:
In section 2, we have reviewed the related works. Section 3 explores the SimSiam algorithm used for pre-training in-domain features from remote sensing images. Section 4 presents the statistics of the selected datasets for each step. In section 5, we solved the downstream tasks and demonstrated the results. Finally, we have concluded the paper in section 6.

## 2. RELATED WORKS

### 2.1. Visual Representation Learning in Remote Sensing Imagery

Where large labeled datasets are unavailable, the general solution is to use pre-trained models on large-scale datasets such as ImageNet. The referred models can be used to extract features from new datasets or as a starting point for fine-tuning weights on othe tasks and datasets. However, this type of transfer learning is directly applicable to the RGB remote sensing datasets[1, 5-12]. For pre-training the in-domain general representations for overhead imagery, we can use either supervised or unsupervised methods. We can refer to [5] as an example of supervised learning-based methods. Here, the initial steps for supervised in-domain visual representation learning from remote sensing images are described. The learned features have been evaluated using fine-tuning on land cover classification datasets. In most cases, it has been demonstrated that in-domain features learned from remote sensing datasets perform better than ImageNet counterparts. Additionally, in the case of supervised learning, they have investigated the characteristics that make a dataset a good reference for learning visual representations. The features learned from multi-resolution datasets have demonstrated higher generalizability and better performance. The researchers in [26] combined the Resisc45, PatternNet, and RSI-CB datasets, trained a model on them, and then fine-tuned it on the UC-Merced dataset. Compared to the ImageNet model, this model is more accurate. In comparison, the analysis in [27] demonstrated that the models pre-trained on the ImageNet perform better than models pre-trained on the PatternNet when transferred to the target AID [28] and UCM [29] datasets. Both studies [5] and [6] are very similar and examined the performance of ImageNet pre-trained models and models pre-trained on the in-domain datasets. They conducted experiments in [6] using two high-resolution and three medium-resolution datasets. The results indicated that fine-tuning in-domain pre-trained weights on remote sensing datasets perform better than ImageNet weights. However, the mentioned work used only two high-resolution datasets for pre-training in-domain features, and the influencial factors on learning highly generalizable representations from remote sensing datasets are not yet fully determined. Additionally, the mentioned works have examined the effect of a pre-training dataset for supervised representation learning. In contrast, by conducting detailed experiments on PatternNet and Resisc45 datasets, we have examined the impact of the pre-training dataset for the SimSiam algorithm, which is a contrastive self-supervised learning method.

### 2.2. A brief overview of self-supervised learning methods

Self-supervised learning is a highly practical subset of unsupervised learning that aims to learn general visual features from images without the use of human labels. In general, self-supervised learning methods consist of two steps. In the first step, a pretext task is designed, and by solving this proxy task, visual representations are learned. The second step is to transfer the pre-trained features from the previous step to other downstream tasks. The resulting model from the first step can be used as a starting point for further fine-tuning or feature extraction. Such techniques will be advantageous for difficult-to-label aerial and medical images. In the following section, we have classified the self-supervised visual representation learning methods into three groups and discussed each one.

**Representation learning by solving pretext tasks:**

Researchers in computer vision have defined many pretext tasks to date. We refer readers to [30] for a review of self-supervised learning methods based on pretext tasks. The authors have classified all pretext tasks into four categories:
1. Generative-based methods such as image-colorization, super-resolution, etc
2. Context-based tasks, such as image jigsaw puzzles, geometric transformations, etc.
3. Free semantic label-based tasks, such as contour detection, depth estimation, etc.
4. Cross Modal-based methods such as optical flow estimation, visual-audio correspondence, etc.

Moreover, the researchers in [31] and [32] combined several pretext tasks for capturing high generalizable features. These studies demonstrated that different pretext tasks are complementary and that combining them results in the acquisition of more generalizable features. The visual representations learned by solving most of the pretext tasks have limited generalizability and performance on downstream tasks compared to the ImageNet pre-trained models.

**Clustering:**

Clustering-based methods are another type of unsupervised method for learning visual representations. They alternate between clustering the representations and learning to predict the cluster assignment. Instead of directly comparing features as in contrastive learning, SwAV[21] clusters the data while enforcing consistency between cluster assignments produced for different augmentations (or views) of the same image. Researchers in [33] demonstrated that k-means assignments could be used as pseudo labels to learn visual representations. In [34], how to cast the pseudo-label assignment problem as an instance of the optimal transport problem have been demonstrated. Despite the fact that clustering-based methods have been very effective for learning visual representations, due to the need to alternate between clustering and feature learning, they require high computing resources.

**Contrastive self-supervised learning**:

The most contrastive self-supervised learning methods have emerged In the last two years. These methods minimized the gap between supervised and unsupervised feature learning. We refer readers to [35] and [36] to learn more about contrastive self-supervised learning algorithms. The main idea of contrastive learning is to bring pairs of positive samples closer together and pairs of negative samples further apart in the feature space. In practice, the performance of contrastive learning methods is highly dependent on a large number of negative samples [35]. As a result, a large number of negative samples must be provided. For instance, the PIRL [37] algorithm stores negative samples in a memory bank, while the MoCo [38] algorithm stores a row of negative samples in a momentum encoder. In contrast, the SimCLR [22] algorithm generates negative examples with large batch sizes, necessitating the use of significant computational resources. Unlike other contrastive self-supervised learning methods, the SimSiam algorithm does not need a memory bank or a larger batch size. Therefore, it requires fewer computation resources.

## 2.3. Self-supervised learning in remote sensing

Recently, some researchers have attempted to apply self-supervised learning algorithms and concepts to remote sensing images. In [16], multi-scale spatial features from high-resolution remote sensing images are captured by multiple-layer feature-matching generative adversarial networks (MARTA GANs) and are used for solving land cover classification tasks. In another work, a pretext task is defined in a way that predicts RGB channels information using high-frequency channels information[18]. Additionally, [15] employs image colorization, relative position prediction, and instance discrimination as pretext tasks for learning the in-domain representations from remote sensing images. The learned features are evaluated by transferring to other land cover classification datasets with very few labeled samples. In [39] the MoCov2 algorithm have been modified by introducing a geography-aware cost function to learn the visual features of remote sensing images. Rather than using regular data augmentation techniques to generate positive samples, the aforementioned work utilized geographic information about the locations that satellites frequently pass through to generate positive samples. Researches in [14] have demonstrated that hierarchical pre-training first on natural images and then on remote sensing images, improves accuracy in downstream tasks. In [40], the effect of different data augmentation methods for contrastive self-supervised learning algorithms on remote sensing datasets has been studied. In [41], a self-supervised learning approach for pre-training weights from remote sensing images is proposed. This approach makes simultaneous use of the correspondence between remote sensing images and geo-tagged audio recordings. Therefore, it pre-trains the features using both the image information and the corresponding audio of each image.

## 3. AN OVERVIEW OF THE SIMSIAM ALGORITHM

The method we used in this paper is divided into two sections:
1. Self-supervised pre-training using the SimSiam algorithm.
2. Transferring the pre-trained weights in the previous step to downstream tasks and evaluating the generalizability of features by fine-tuning on various land cover classification datasets.

The features learned using most contrastive learning algorithms are highly generalizable. However, the main drawback of these methods is their high computational requirements. In general, this requirement for plenty of computational resources is motivated by the following three main items: 1. negative samples; 2. large batch size; 3. momentum encoder. Unlike other contrastive learning methods, the SimSiam algorithm does not require any of the three items mentioned above, making it significantly more computational resource-efficient. Therefore, we employ this algorithm to learn the visual features of remote sensing images [24]. The schematic of this algorithm is shown in Figure 1:

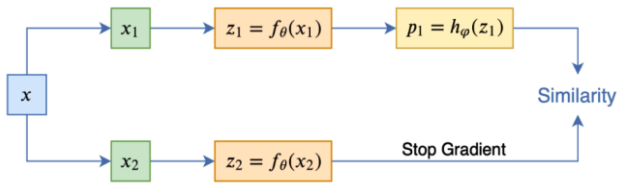

**Figure 1**: SimSiam architecture

After applying different data augmentation techniques to the image x, two distinct views, $x_1$ and $x_2$, are generated. Both of these two distinct perspectives are entered into both sides of the Siamese architecture and then processed using the encoder f. This encoder utilizes a ResNet50 backbone, and an MLP projection head to extract the features from the input image. Therefore, each image is converted from pixel space to a smaller feature space. The projection head in the encoder f consists of three layers of MLP with batch normalization layers applied to each fully connected layer, including the output layer. Following the encoder(f), the architecture only has a top-side prediction head module. The prediction head MLP(h) merges the encoder outputs for the $x_1$ view and then matches its dimensions to the encoder output on the bottom side of the architecture. This MLP (h) is composed of two fully connected layers and a hidden layer that has been subjected to a batch normalization layer. The trained weights are shared on both sides of the model. [28] demonstrated that copying weights on both sides of Siamese architectures produces poor results. They have provided a momentum update to avoid this issue. The proposed solution has the disadvantage of requiring a large number of computational resources. The SimSiam employs a stop-gradient operator on one side of the architecture to overcome the requirement for high computational resources as the BYOL[23] algorithm does. When the stop-gradient operator is applied to any side of the network, the gradients on that side are not updated via backpropagation. The proposed cost function is simple and is defined as a function of the cosine similarity of two vectors. If we illustrate the output vectors of two views with $p_1 \triangleq h(f(x_1))$ and $z_2 \triangleq f(x_2)$, where $h(f(x_1))$ represents the output of the prediction head applied to $f(x_1)$ and $f(.)$ is the backbone which is applied to both views, then the objective function can be defined as follows:

$$\mathcal{D}(p_1, z_2) = -\frac{p_1}{\|p_1\|_2} \cdot \frac{z_2}{\|z_2\|_2}$$

In this equation, $\|.\|$ represents the $L_2$ norm. Finally, the total cost function is a symmetric function, which is defined as follows:

$$L = \frac{1}{2}\mathcal{D}(p_1, z_2) + \frac{1}{2}\mathcal{D}(p_2, z_1), p_2 \triangleq h(f(x_2))$$

The cost obtained for all images in the batch is averaged and considered as the total loss. The stop gradient operator is a critical component that makes this algorithm work well. This operator applies to the features extracted from each view; therefore, the final cost function is defined as follows:

$$L = \frac{1}{2}\mathcal{D}(p_1, stopgrad(z_2)) + \frac{1}{2}\mathcal{D}(p_2, stopgrad(z_1))$$

This relationship demonstrates that the defined cost function is perfectly symmetric. Additionally, gradients are updated only when a corresponding view enters the network from the top side of the architecture [24].

## 4. DATASETS

We conducted our experiments using two sets of remote sensing datasets. The first category contains datasets selected for self-supervised pre-training with the SimSiam algorithm, while the second category contains datasets used to evaluate the features learned.

### 4.1. Self-supervised pre-training datasets

We have used MLRSNet, NWPU-RESISC45, and PatternNet, to pre-train the general representations using the SimSiam. These datasets have different characteristics from which the number of classes, number of samples, and spatial resolution are more noticeable discrepancies. The variety of attributes in pre-train datasets can help us to identify the vital factors that make the dataset a good choice for acquiring general representations.

**MLRSNet:** MLRSNet is a multi-label high spatial resolution remote sensing dataset for semantic scene understanding. It contains 109,161 remote sensing images that are annotated into 46 categories, and the number of sample images in a category varies from 1,500 to 3,000. The images have a fixed size of 256×256 pixels with various pixel resolutions (~0.1m to 10m). Moreover, the number of labels associated with each image varies from 1 to 13. The dataset can be used for multi-label based image classification, multi-label based image retrieval, and image segmentation.

**NWPU-RESISC45[10]:** This dataset contains 31.5k images classified into 45 classes. The images of this dataset have a high spatial resolution. For many samples, this dataset has a spatial resolution of between 0.2m to 30m per pixel. During the pre-training phase, we used all of the images in the dataset. During the transfer learning phase and fine-tuning of learned weights, we used 60% of the data for training, 20% for validation, and 20% for testing.

**PatternNet[25]:** PatternNet has a higher spatial resolution than Resisc45 and consists of 38 classes with 800 images per class. Therefore, there are 30.4k samples in this dataset. The image size of this dataset is 256x256. Additionally, the spatial resolution of this dataset is between 0.06m to 4.96m.

### 4.2. Downstream Datasets

In addition to the Resisc45 and PatternNet, we have evaluated the pre-trained representations using three distinct datasets with the following characteristics:

**AID[28]**: The dataset contains 10,000 RGB images with a resolution of 600x600 pixels divided into 30 classes. The spatial resolution of images is about 0.5m to 8m.

**EuroSAT[9]**: The dataset contains 27,000 images with 64x64 pixel dimensions classified into ten classes. This dataset has two versions of 13 channels and three RGB channels. The spatial resolution of each image in this dataset is about 10m to 30m, indicating that it has a low spatial resolution. We conducted our experiments using a three-channel version.

**UC_Merced[29]:** The dataset has 2,100 images divided into 21 classes with a resolution of 0.3m and image sizes 256x256.

The following table summarizes the characteristics of the datasets used in this article.

**Table I**: General characteristics of the selected datasets.

| Dataset | Image size | Size | Classes | Resolution (m) |
|---------|------------|------|---------|----------------|
| MLRSNet | 256x256 | 109.16k | 46 | 0.1-10m |
| Resisc45 | 256x256 | 31.5k | 45 | 0.2-30m |
| PatternNet | 256x256 | 30.4k | 38 | 0.06-4.96m |
| AID | 600x600 | 10k | 30 | 0.5-8m |
| UCM | 256x256 | 2.1k | 21 | 0.3m |
| EuroSat | 64x64 | 27k | 10 | 10-30m |

## 5. EXPERIMENTS

Our primary objective is to obtain meaningful representations from aerial imagery in an unsupervised manner and use them to tackle the domain difference issue. As a result, we use the obtained weights either as initial weights or feature extractors. Some of our experiments are inspired by [17]. However, they considered the supervised approach to learning visual representations. Additionally, through extensive experiments, we examine the effect of the pre-training dataset using the SimSiam algorithm. We have conducted our experiments using the PyTorch and PyTorch Lightning[42] frameworks on an Ubuntu system equipped with QuadroP6000 GPU. We repeated each experiment five times and reported the average results.

### 5.1. Self-supervised pre-training using SimSiam

In the first phase, we performed in-domain self-supervised pre-training using the SimSiam algorithm on all instances of the MLRSNet, Resisc45, and PatternNet datasets. As previously described, the SimSiam algorithm utilizes an encoder (f) that consists of a backbone and a projection head. In our experiments, we used ResNet50 as the backbone and applied slight changes to the number of neurons in the projection and prediction head. The projection head is a three-layer MLP with 1024 and 512 neurons in its hidden layers. The predictor module (h) follows the encoder module. It consists of a two-layer MLP with 256 neurons in its hidden layer. We trained the obtained SimSiam on MLRSNet, Resisc45, and PatternNet, for 100k iterations. We used the SGD optimizer with a batch size of 128 and a base learning rate of 0.05 during training, as well as the MultiStepLR scheduler. We set the weight decay and SGD momentum to $10^{-5}$ and 0.9, respectively. We also used ImageNet pre-trained weights during self-supervised pre-training. According to the results of [14], it leads to the higher accuracy of downstream tasks and decreases the time required for convergence. By performing this experiment, we obtained three distinct models pre-trained on datasets with different characteristics.

### 5.2. Transfer Learning to downstream tasks

In this experiment, we fine-tuned the resulting models on five remote sensing datasets with different characteristics and reported global accuracy for each dataset to evaluate the quality of pre-trained representations. We used 60% of the datasets as a training set, 20% as a validation set, and the remaining 20% as a test set to solve downstream tasks. We also used the Adam optimizer with a batch size of 64 and the ReduceLrOnPlateau scheduler. We have fine-tuned all of the models for 100 epochs.

Our data augmentation pipelines are as follows:

We first resize all images to 256x256 pixels for all downstream datasets except EuroSAT and then apply random horizontal or vertical flips. We crop 224x224 pixels from the center of the resulting image. Finally, each dataset is normalized using the mean and standard deviation of the pixel intensities. We repeated each experiment five times and reported the average results.

In TableII, we compared our results to those reported in [17].

**TableII**: Accuracy over different training methods. Our best results were obtained by fine-tuning in-domain representations captured by the SimSiam algorith (Averaged over 5 runs).

| Target dataset → Method ↓ | UCM | EuroSAT | Resisc45 |
|---------------------------|------|---------|----------|
| Scratch[17] | 95.7 | 98.5 | 95.5 |
| ImageNet[17] | 99.2 | 99.1 | 96.6 |
| Supervised In-Domain[17] | 99.6 | 99.2 | 96.8 |
| **Ours** | **99.9** | **99.3** | **97.2** |

In [17], the ResNet50 model is pre-trained in a supervised manner on various remote sensing datasets. The model's final parameters are then fine-tuned using the Resisc45, UCM, Eurostat, and other datasets. As shown in TableII, the self-supervised pre-trained model on the high-resolution

PatternNet dataset outperforms ImageNet pre-trained model and other in-domain supervised models.
In Table III, we have compared our best results obtained by Sim-PatternNet to some of the best available models. The results indicate that self-supervised pre-training using the SimSiam algorithm produced the best results across different land cover classification datasets.

**TableIII**: Comparison of results on selected remote sensing datasets. Our best results were obtained by fine-tuning in-domain representations captured by the SimSiam algorith (Averaged over 5 runs).

| Dataset | Reference | Description | Acc (%) |
|---|---|---|---|
| AID | [15] | Unsupervised | 78.83 |
|  | [41] | Unsupervised | 84.44 |
|  | [9] | Supervised | 94.38 |
|  | [19] | Supervised | 95.58 |
|  | [6] | Supervised | 97.30 |
|  | **Ours** | **Unsupervised** | **97.83** |
| EuroSAT | [15] | Unsupervised | 76.37 |
|  | [5] | Supervised | 99.20 |
|  | [43] | Unsupervised | 98.91 |
|  | [9] | Supervised | 98.57 |
|  | **Ours** | **Unsupervised** | **99.26** |
| UCM | [41] | Unsupervised | 89.71 |
|  | [9] | Supervised | 96.42 |
|  | [5] | Supervised | 99.61 |
|  | [44] | Supervised | 92.40 |
|  | [45] | Supervised | 97.10 |
|  | [12] | Supervised | 99.41 |
|  | [7] | Supervised | 98.50 |
|  | **Ours** | **Unsupervised** | **99.90** |
| PatternNet | [25] | Supervised | 96.65 |
|  | [6] | Supervised | 99.84 |
|  | **Ours** | **Unsupervised** | **99.90** |
| Resisc45 | [41] | Unsupervised | 84.88 |
|  | [43] | Unsupervised | 96.28 |
|  | [5] | Supervised | 96.83 |
|  | [6] | Supervised | 97.03 |
|  | **Ours** | **Unsupervised** | **97.20** |

## 5.3. Choosing the appropriate dataset for self-supervised pre-training

### A. Fine-tunning all layers

In this section, by conducting detailed experiments, we have examined the effect of the pre-training dataset on the final accuracy of downstream tasks to determine the effective characteristics for selecting the pre-training dataset using the SimSiam algorithm. For this purpose, we have used MLRSNet, Resisc45, and PatternNet for pre-training using the SimSiam. These datasets have different attributes in the number of samples, class diversity, and spatial resolutions. The similarity of the pre-training or source dataset to the downstream or target dataset is a critical factor affecting the accuracy of land cover classification tasks, as has been discussed for the supervised approach [17]. However, for representation learning from remote sensing images using a contrastive self-supervised learning approach, potentially influential factors must be examined through deliberate experiments.

**Table IV**: Class similarity of MLRSNet, Resisc45 and PatternNet to the downstream tasks. PD and DS stand for Pretraining Dataset and DownStream task, respectively

| DS → <br> PD↓ | AID | EuroSAT | UCM |
|---|---|---|---|
| MLRSNet | 66.6% | 40% | 76.1% |
| PatternNet | 30% | 30% | 85.71% |
| Resisc45 | 60% | 40% | 90.47% |

The class similarity is a proxy that shows the similarity of source and target datasets. We calculated the similarities by comparing the number of identical classes in both pre-training and downstream datasets. Table 4 indicates that the downstream datasets used in our experiment are, on average, more similar to Resisc45 than PatternNet and MLRSNet. Another factor that causes the learning of global features and, as a result, high performance in target datasets is the class diversity of the pre-training dataset [17]; the higher the class diversity of the source dataset, the higher the generalizability of the pre-trained features on target tasks. Although Resisc45 has higher class diversity and more similarity to the target datasets, the pre-trained features on the PatternNet have a higher generalization ability on all downstream tasks. A vital factor that comes into view is the spatial resolution of source datasets. It is between 0.06m and 4.96m for PatternNet, 0.1m to 10m for MLRSNet, and 0.2m and 30m for Resisc45. Therefore, the spatial resolution of PatternNet is higher than Resisc45 and MLRSNet. The high spatial resolution in remote sensing images makes the edges of the objects in the images sharper, and because self-supervised learning methods provide supervisory signals from the data, the presence of these edges makes the differences between the objects in the images more accentuated. As a result, the SimSiam model can better learn the difference between the objects in the dataset. It is conclusive that the importance of other factors such as class similarity, class diversity, and the number of samples for learning general features from remote sensing images using the SimSiam are highlighted when the pre-training dataset has a high spatial resolution. MLRSNet is much larger than PatternNet, but the generalizability of PatternNet is better than that of MLRSNet. It means that; although the class diversity, class similarity, and number of samples of the PatternNet dataset to the target datasets are lower than Resisc45 and MLRSNet, these factors are still significant for the PatternNet dataset. All these factors make PatternNet an appropriate source for pre-training visual features using the SimSiam algorithm.

We compared the results of fine-tuning pre-trained weights on the PatternNet and Resisc45 datasets in Table 5. These results demonstrate that despite Resisc45's high similarity to downstream datasets and the diversity of its classes, the pre-trained model on the PatternNet performs significantly better when solving land cover classification tasks.
We compared the results of fine-tuning different pre-trained models in Table 5.

**TableV**: Results on downstream Tasks (Averaged over 5 runs). The pre-trained model on the PatternNet performs better than other models. PD and DS stand for Pretraining Dataset and DownStream task, respectively.

| PD → <br><br> DT ↓ | AID | EuroSAT | UCM |
|---|---|---|---|
| Resisc45 | 97.62 | 97.75 | 98.24 |
| MLRSNet | 97.78 | 98.45 | 98.85 |
| PatternNet | **97.83** | **99.26** | **99.90** |

According to Table 4, the similarity of the AID to the MLRSNet, Resisc45, and PatternNet is 66.6%, 60%, and 30%, respectively. However, fine-tuning weights obtained from the PatternNet dataset on the AID perform better than other models. These results indicate that when using the SimSiam algorithm to train visual representations from remote sensing images, the choice of higher-resolution datasets is a critical factor with a huge impact on the final performance of downstream tasks. However, the conclusions are made based only on three datasets, and additional experiments with diverse datasets are required to make more precise generalizations.

## B. Linear Evaluation with limited number of samples

In this section, by using linear evaluation, we further examined the quality of the pre-trained features. In the figure below, we have shown the general outline of the linear evaluation.

**Figure 1**: Chematic of a linear evaluator. Pre-trained model serves as feature extractor.

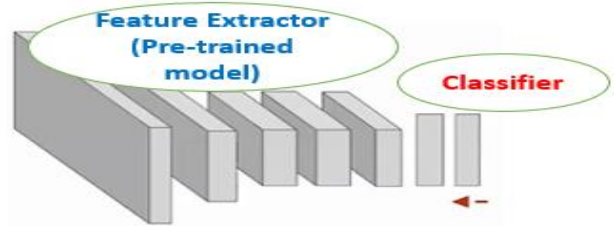

For linear evaluation of pre-trained features, we freeze the backbone of the ResNet50 model and only train the head, which is the classifier. Therefore, the backbone of the model serves as the feature extractor. Since fine-tuning the whole model distorts all the pre-trained weights, we believe it can not provide an ideal solution for the evaluation of different pre-trained models. In contrast, fine-tuning the last layer does not change the pre-trained weights. Therefore, it better shows the capability of the pre-trained backbone in extracting general features from remote sensing images. Using a limited number of samples per class, we conducted our experiments and reported the global accuracy in Table VI.

According to Tabel VI, all the in-domain pre-trained models using the SimSiam algorithm on remote sensing datasets perform much better than the ImageNet model, and the smaller the number of training samples, The more discrepancy in the performance of pre-trained models becomes evident. As in the previous experiments, in linear evaluation, the PatternNet pre-trained model performs better than other models in most cases. The acquired results lead us to use unsupervised models for capturing in-domain features from remote sensing images. One critical point that should be carefully considered is the characteristics of the source dataset. The results showed that the source dataset not only should have high-class diversity, an enormous number of samples, and high similarity with target datasets but also should have a high spatial resolution.

**TableVI**: Results of linear evaluation under limited number of samples (Averaged over 5 runs). The pre-trained model on the PatternNet performs better than other models. PD and DS stand for Pretraining Dataset and DownStream task, respectively.

| DT → <br><br> PD ↓ | AID | | | | EuroSat | | | | UCM | | | |
|---|---|---|---|---|---|---|---|---|---|---|---|---|
| | Number of images per class | | | | Number of images per class | | | | Number of images per class | | | |
| | 5 | 10 | 20 | 50 | 5 | 10 | 20 | 50 | 5 | 10 | 20 | 50 |
| ImageNet | 45.45 | 52.36 | 63.14 | 70.17 | 39.36 | 46.45 | 51.22 | 59.71 | 40.43 | 50.33 | 56.72 | 63.21 |
| Resisc45 | 72.32 | 75.44 | 81.74 | 86.56 | 77.50 | 80.12 | 85.16 | 90.93 | 77.89 | 82.11 | 87.95 | 92.15 |
| MLRSNet | 73.34 | 77.10 | 82.52 | **89.52** | 79.31 | 83.27 | 88.87 | 92.58 | 80.92 | 84.60 | 90.37 | **94.85** |
| **PatternNet** | **73.89** | **78.25** | **85.13** | 89.33 | **80.02** | **84.19** | **89.55** | 92.31 | **81.65** | **85.87** | **91.70** | 94.66 |

## 6. CONCLUSIONS

Recently, contrastive learning, a subset of self-supervised learning, has made significant progress in general visual representations of natural images. The available remote-sensing datasets have different numbers of samples and channels, spatial resolutions, and image sizes. Therefore, it is necessary to examine the transferability of self-supervised pre-trained features from remote sensing images and determine the right factors that make the dataset a good choice for feature pre-training. In this paper, we utilized the SimSiam for in-domain general feature learning from three remote-sensing datasets with different characteristics. The pre-trained weights were then evaluated by fine-tuning and linear evaluation on other land cover classification datasets achieving state-of-the-art results. Our deliberate experiments demonstrate that for contrastive self-supervised pre-training of remote-sensing images, higher resolution datasets lead to better performance on downstream tasks.

**Data Availability Statement (DAS)**

The data that support the findings of this study are available from the corresponding author, [H.S.], upon reasonable reques.